\documentclass{article}

\usepackage[main, preprint]{mystyle}

\usepackage[utf8]{inputenc} % allow utf-8 input
\usepackage[T1]{fontenc}    % use 8-bit T1 fonts
\usepackage{natbib}
\usepackage{url}            % simple URL typesetting
\usepackage{booktabs}       % professional-quality tables
\usepackage{amsfonts}       % blackboard math symbols
\usepackage{nicefrac}       % compact symbols for 1/2, etc.
\usepackage{microtype}      % microtypography
\usepackage{xcolor}         % colors
\usepackage{subcaption}
\usepackage{setspace}
\usepackage{graphicx}
\usepackage{multirow}
\usepackage{amsmath}
\usepackage{adjustbox}
\usepackage{caption}
\usepackage{enumitem}
\usepackage{makecell}
\usepackage{colortbl}
\usepackage{wrapfig}
\usepackage{pifont}
\usepackage[most]{tcolorbox}
\usepackage{algorithm}          % For algorithm float environment
\usepackage{listings}           % For code-style pseudocode
\usepackage{marvosym}
\usepackage{bxcoloremoji}
\setlist[itemize]{leftmargin=*}
\definecolor{mydarkblue}{rgb}{0,0.08,0.45}

\usepackage{amsmath,amsfonts,bm}

\def\eqref#1{equation~\ref{#1}}
\def\1{\bm{1}}

\DeclareMathAlphabet{\mathsfit}{\encodingdefault}{\sfdefault}{m}{sl}
\SetMathAlphabet{\mathsfit}{bold}{\encodingdefault}{\sfdefault}{bx}{n}

\usepackage{graphicx}
\usepackage{booktabs}
\usepackage{amssymb}
\usepackage{stmaryrd}
\usepackage[table]{xcolor}
\usepackage{pifont}

\definecolor{mygray}{HTML}{808080}
\definecolor{mydarkgreen}{HTML}{1B7F3A}

\usepackage{listings}
\usepackage{hyperref}
\usepackage{url}
\usepackage{amsmath,amssymb,mathtools}
\usepackage{amsthm}
\usepackage{enumitem}
\usepackage{booktabs}
\usepackage{algpseudocode} % algorithmicx

\usepackage[table]{xcolor}
\usepackage{array}
\newcolumntype{g}{>{\columncolor{gray!10}}c}
\usepackage{multirow}           
\usepackage{adjustbox}
\usepackage{hyperref}
\usepackage{url}
\usepackage{subcaption}
\usepackage{booktabs} 
\usepackage{array}
\usepackage{algorithm}
\usepackage{amsmath}
\definecolor{step1color}{HTML}{D5E8D4} % Light Green
\definecolor{step2color}{HTML}{A8D08D} % Medium Green
\definecolor{step3color}{HTML}{82B366} % Darker Green
\definecolor{step4color}{HTML}{5A8A42} % Even Darker Green
\definecolor{promptcolor}{HTML}{F5F5F5} % Light Gray for prompt background
\definecolor{cond}{HTML}{2E75B6}
\definecolor{hdpocolor}{HTML}{2E75B6}
\definecolor{metiscolor}{HTML}{F3B000}

\newcommand{\hdpo}{\textcolor{hdpocolor}{\textbf{HDPO}}}
\newcommand{\metis}{\textcolor{metiscolor}{\textbf{Metis}}}

\title{Act Wisely: Cultivating Meta-Cognitive Tool Use in Agentic Multimodal Models}

\author{
	\textbf{Shilin Yan}$^{1*\dagger}$\textsuperscript{\ddag} \quad
    \textbf{Jintao Tong}$^{1,2*}$ \quad 
    \textbf{Hongwei Xue}$^{1\dagger}$\quad 
    \textbf{Xiaojun Tang}$^{1}$\quad
    \textbf{Yangyang Wang}$^{1}$ \and
    \textbf{Kunyu Shi}$^{1}$\quad  
    \textbf{Guannan Zhang}$^{1}$\quad 
    \textbf{Ruixuan Li}$^{2}$\textsuperscript{\ddag} 
    \quad \textbf{Yixiong Zou}$^{2}$\textsuperscript{\ddag}
	\and \\	
	$^1$Accio Team, Alibaba Group~~
	$^2$Huazhong University of Science and Technology \and
     $^{\dagger}$Project Leader \quad \textsuperscript{\ddag}Corresponding Author
}

\headerlogos{
	\includegraphics[height=0.5cm]{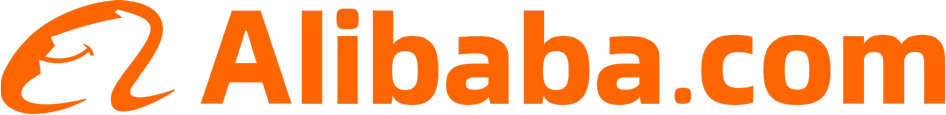}\hspace{0.3cm}
	\includegraphics[height=0.5cm]{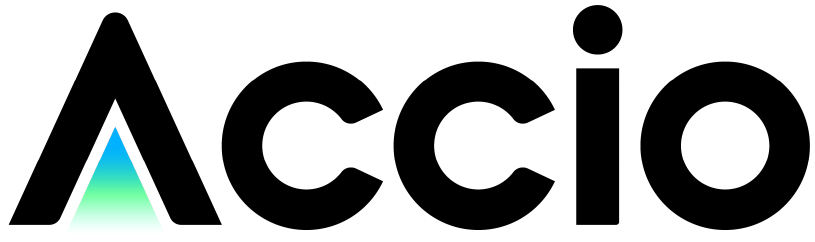}  %\hfill
}

\newcommand{\github}{\raisebox{-1.5pt}{\includegraphics[height=1.05em]{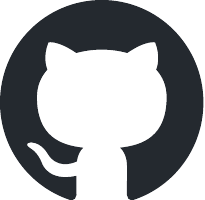}}}
\begin{document}
	
\maketitle

\insert\footins{\noindent\footnotesize $^*$Equal contribution. This work was done during Jintao Tong's internship at the Accio Team, Alibaba Group.}

\vspace{-0.5cm}
\begin{abstract}
The advent of agentic multimodal models has empowered systems to actively interact with external environments. However, current agents suffer from a profound meta-cognitive deficit: they struggle to arbitrate between leveraging internal knowledge and querying external utilities. Consequently, they frequently fall prey to blind tool invocation, resorting to reflexive tool execution even when queries are resolvable from the raw visual context. This pathological behavior precipitates severe latency bottlenecks and injects extraneous noise that derails sound reasoning. Existing reinforcement learning protocols attempt to mitigate this via a scalarized reward that penalizes tool usage. Yet, this coupled formulation creates an irreconcilable optimization dilemma: an aggressive penalty suppresses essential tool use, whereas a mild penalty is entirely subsumed by the variance of the accuracy reward during advantage normalization, rendering it impotent against tool overuse. To transcend this bottleneck, we propose \hdpo, a framework that reframes tool efficiency from a competing scalar objective to a strictly conditional one. By eschewing reward scalarization, \hdpo{} maintains two orthogonal optimization channels: an accuracy channel that maximizes task correctness, and an efficiency channel that enforces execution economy exclusively within accurate trajectories via conditional advantage estimation. This decoupled architecture naturally induces a cognitive curriculum—compelling the agent to first master task resolution before refining its self-reliance. Extensive evaluations demonstrate that our resulting model, \metis, reduces tool invocations by orders of magnitude (e.g., from 98\% to 2\%) while simultaneously elevating reasoning accuracy. By shattering the illusion that heavy tool reliance equates to better performance, \metis{} pioneers a shift from merely executing tools to cultivating the meta-cognitive wisdom of abstention. \\

\coloremojicode{1F310}  \textbf{Project Page:} \href{https://Accio-Lab.github.io/Metis}{https://Accio-Lab.github.io/Metis} \vspace{0.1cm}

\github{}  \textbf{Github Repo:} \href{https://github.com/Accio-Lab/Metis}{https://github.com/Accio-Lab/Metis} \vspace{0.1cm}

\coloremojicode{1F917}  \textbf{HuggingFace:} \href{https://huggingface.co/Accio-Lab/Metis-8B-RL}{https://huggingface.co/Accio-Lab/Metis-8B-RL}
	
\end{abstract}

\section{Introduction}

\vspace{0.1cm}
\begin{quote}
\textit{``The art of being wise is the art of knowing what to overlook.''} \\
\hspace*{0pt}\hfill --- William James
\end{quote}
\vspace{0.2cm}

\begin{figure*}[!t]
		\centering
		\includegraphics[width=\linewidth]{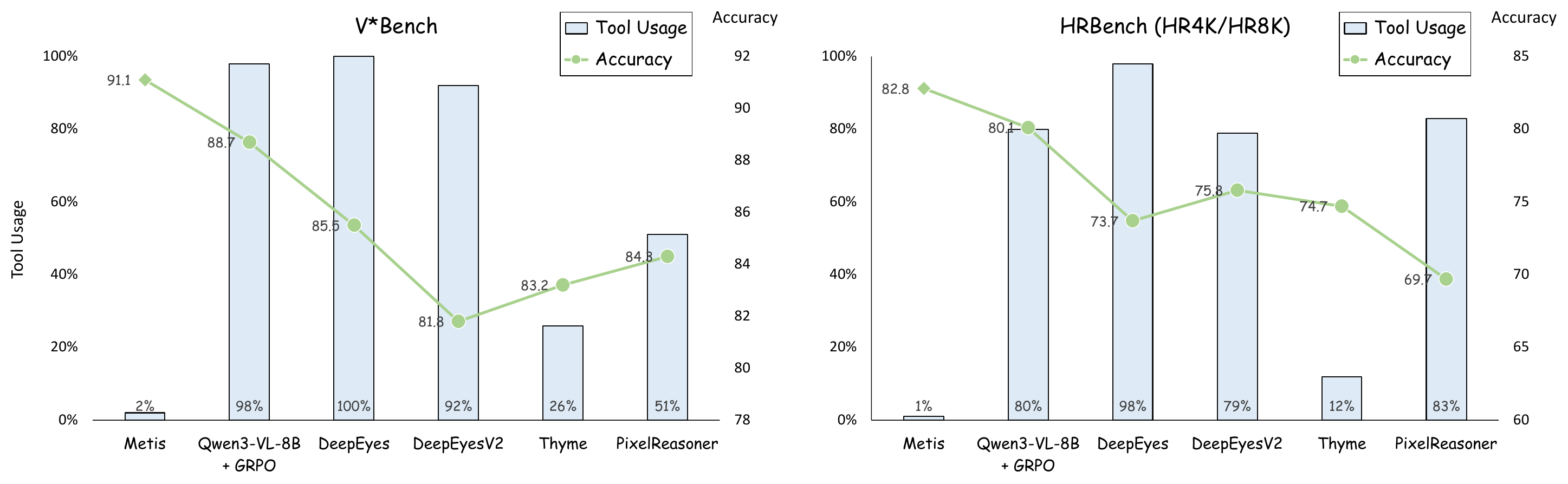} 
		\caption{\textbf{Comparison of tool-use efficiency and task performance.} Existing methods rely heavily on tool calls, reflecting limited efficiency awareness. In contrast, our method uses tools far more selectively while achieving the best overall performance, showing that strong accuracy and high efficiency can be attained simultaneously.}
		\label{Fig.intro_analysis}
	\end{figure*}

The evolution of multimodal large language models (MLLMs)~\citep{liu2023visual,hurst2024gpt,liu2024llavanext,bai2025qwen2,wang2025internvl3,bai2025qwen3,comanici2025gemini,gemini3_1} into autonomous, agentic systems has catalyzed a new paradigm in complex visual reasoning. By interleaving internal cognitive processes with active environmental interactions~\cite{wang2025pixel,zhang2025thyme,zheng2025deepeyes}, these multimodal agents can dynamically acquire fine-grained visual evidence, execute intermediate computations, and transcend the inherent limitations of static parametric knowledge. This approach has yielded substantial progress across diverse domains, including visual question answering, document understanding, and multi-step decision making~\cite{qiao2025we,wang2024charxiv,yue2024mmmu}.

Despite these expanded capabilities, current agents suffer from a profound meta-cognitive deficit: they struggle to dynamically arbitrate between leveraging internal parametric knowledge and querying external utilities. Discerning the genuine necessity of a tool requires the agent to calibrate its own epistemic uncertainty against the sufficiency of the visual context—a sophisticated meta-cognitive skill notoriously difficult to instill via standard supervised fine-tuning. Without such calibration, state-of-the-art open-source agents~\cite{hong2025deepeyesv2,wu2024v,zheng2025deepeyes} frequently fall prey to \textbf{blind tool invocation}, resorting to reflexive tool execution even when queries are intrinsically resolvable from the raw visual input. As empirically demonstrated in Figure~\ref{Fig.intro_analysis}, existing models exhibit a stark imbalance: they incur exorbitant tool invocation rates (frequently exceeding 80\% to 90\%), yet fail to translate this computational expenditure into superior reasoning performance. This pathological behavior is highly detrimental. Prevailing reinforcement learning paradigms exhibit a myopic focus on task completion, engendering \textbf{latency-agnostic optimization}. In real-world agentic deployments, the solution space for a given query encompasses a multitude of valid trajectories. Yet, owing to the serial bottleneck inherent in external API invocations, these trajectories diverge profoundly in their temporal footprint. Without explicit optimization for execution economy, models inevitably degenerate into functionally competent but practically sluggish systems. Furthermore, redundant tool interactions inject extraneous environmental noise that frequently derails otherwise sound reasoning trajectories and degrades final performance.

\begin{figure*}[!t]
	\centering
	\includegraphics[width=\linewidth]{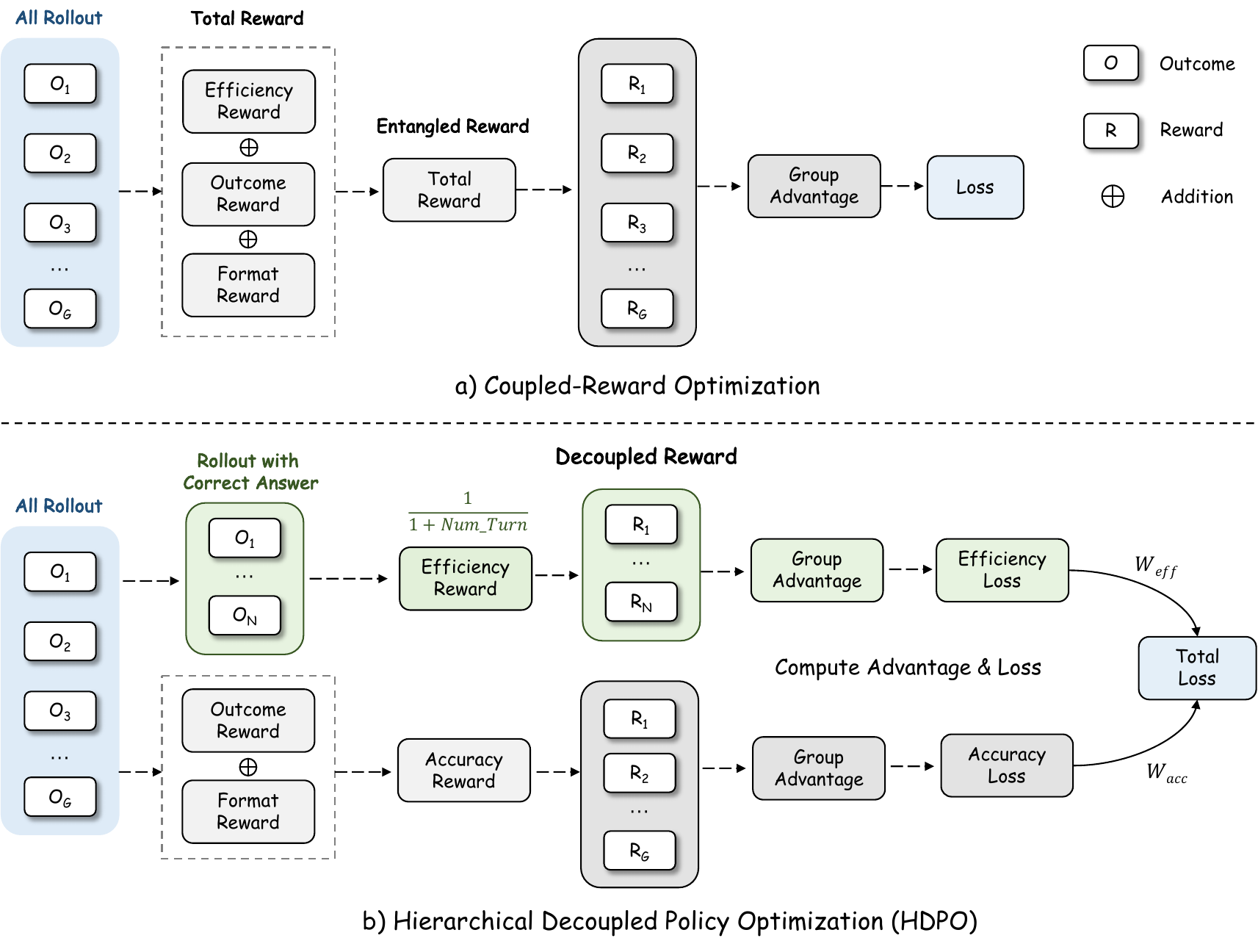} 
	\caption{\textbf{Comparison between coupled-reward optimization and \hdpo{}.} Existing methods entangle accuracy and efficiency into a single reward signal, while \hdpo{} decouples them into separate branches and combines them only at the final loss, enabling more strategic tool use.}
	\label{Fig.intro}
\end{figure*} 

A prevalent mitigation strategy is to penalize excessive tool usage during reinforcement learning (RL). However, as illustrated in Figure~\ref{Fig.intro}(top), existing protocols~\cite{song2025codedance,wang2025adatooler} typically scalarize task accuracy and tool efficiency into a singular reward formulation. This \textbf{coupled design} precipitates an irreconcilable optimization dilemma. An aggressive efficiency penalty renders the model overly conservative, suppressing essential tool use on arduous tasks and thereby sacrificing correctness. Conversely, a mild penalty is entirely subsumed by the variance of the accuracy reward during advantage normalization (e.g., in GRPO). For instance, an inaccurate trajectory with zero tool calls might yield a mixed reward mathematically indistinguishable from an accurate trajectory with excessive tool usage, severely confounding the policy gradient. Consequently, the efficiency signal is effectively \textbf{``washed out,''} rendering the penalty impotent against tool overuse on simpler tasks. A scalarized reward is thus fundamentally inadequate for fostering the instance-dependent, strategic arbitration required for intelligent tool use.

To transcend this bottleneck, we propose \textbf{Hierarchical Decoupled Policy Optimization (\hdpo{})}, an RL framework that reframes tool efficiency from a competing scalar objective to a strictly conditional one. As shown in Figure~\ref{Fig.intro}(bottom), \hdpo{} eschews the mixed reward. Instead, it maintains an \emph{accuracy channel} that globally maximizes task correctness across all rollouts, and an \emph{efficiency channel} that enforces tool parsimony exclusively within accurate trajectories via a novel {conditional advantage mechanism}. By decoupling these objectives until the final loss computation, \hdpo{} eliminates gradient interference and establishes a natural cognitive curriculum: compelling the agent to first master task resolution before refining its self-reliance. Crucially, recognizing that strategic RL requires a high-fidelity environment, we complement \hdpo{} with a rigorous data curation pipeline to eradicate hallucinated environmental dynamics and isolate genuine tool necessity.

Inspired by the principle of parsimony, we train a strategic multimodal reasoning agent, \textbf{\metis{}}, equipped with coding and searching tools. Rather than treating tool invocation as a default reflex, the agent learns to use tools only when they provide genuinely useful evidence or computation. As shown in Figure~\ref{Fig.intro_analysis}, our approach shatters the conventional reliance on heavy tool usage, achieving state-of-the-art accuracy with near-zero redundant tool invocations (e.g., 2\% vs. 98\% for standard GRPO). Our results demonstrate that strategic tool use and strong reasoning performance are not a trade-off; rather, eliminating noisy, redundant tool calls directly contributes to superior accuracy. More broadly, our work suggests a paradigm shift in tool-augmented learning: from merely teaching models \emph{how} to execute tools, to cultivating the meta-cognitive wisdom of \emph{when} to abstain from them. In summary, this work makes the following contributions:

\vspace{-0.3cm}
\begin{itemize}[leftmargin=*]
	\setlength{\itemsep}{1pt}
	\setlength{\parskip}{1pt}
	\item \textbf{Problem Formulation.} We identify blind tool invocation as a critical pathological behavior in multimodal agents and expose the mathematical and semantic vulnerabilities of coupled-reward RL, demonstrating how efficiency signals are systematically subsumed by accuracy variance.
	
	\item \textbf{Algorithm.} We propose \hdpo{}, a framework that eschews reward scalarization to provide clean, orthogonal learning signals. By introducing a conditional advantage formulation, \hdpo{} ensures that tool parsimony is optimized exclusively within accurate trajectories, compelling the agent to prioritize correctness before efficiency.

	\item \textbf{Model \& Performance.} We train \metis{}, a strategic multimodal agent that achieves state-of-the-art performance across diverse benchmarks. By reducing tool usage by over 90\% while simultaneously elevating reasoning accuracy, our results empirically validate that true execution efficiency acts as a catalyst for, rather than a trade-off against, superior reasoning performance.
	
\end{itemize}

\section{Related Works}

\subsection{Multimodal Large Language Models.}
Multimodal large language models (MLLMs)~\citep{bai2025qwen2,liu2024llavanext,wang2025internvl3,bai2025qwen3, yan2025crosslmm} have achieved strong performance on a wide range of vision-language tasks by integrating visual encoders with large language models~\cite{bai2023qwen,liu2023visual}. Early MLLMs mainly focus on direct answer generation for tasks such as visual question answering and image understanding~\cite{liu2024llavanext,li2024llava, wang2024qwen2}. Inspired by the success of chain-of-thought in LLMs, recent MLLMs introduce explicit intermediate reasoning to handle more complex multimodal problems~\cite{kojima2022large,wei2022chain}. These models generate step-by-step textual rationales before producing final answers, leading to improvements on complex multimodal reasoning tasks~\cite{tong2025emosync,xu2025llava,yu2025perception,zhang2025openmmreasoner}. More recently, several works explore latent visual reasoning~\cite{li2025latent,tong2025sketch,tong2026swimbird} by inserting continuous visual representations into the reasoning process, which further improves spatial reasoning ability~\cite{zhang2026think3d}. 
However, despite these advances, most existing MLLMs~\cite{liu2025visual,shen2025vlm} remain passive in that they mainly interpret inputs and generate responses, without actively invoking external tools for retrieval or computation, which limits their reliability on challenging reasoning tasks.

\subsection{Agentic Multimodal Models.}
A growing line of research equips MLLMs with agentic capabilities, allowing them to invoke external tools and interact with the environment during inference rather than relying solely on one-shot prediction~\cite{yao2022react,zhang2025thyme,zheng2025deepeyes}. In the multimodal setting, these tools often include visual operations such as cropping, grounding, image search and so on~\cite{jin2025search,wang2025pixel,wu2025mmsearch}. Such agents have shown strong performance on challenging tasks that require detailed inspection, iterative evidence gathering, or intermediate computation, especially when the raw visual input alone is insufficient~\cite{su2025thinking}.

Despite these advantages, agentic multimodal models also introduce a larger decision space. The model must not only reason about the task itself, but also decide whether to call a tool, which tool to use, and how to incorporate returned observations into subsequent reasoning. Existing work has largely emphasized stronger tool capabilities and better multi-step interaction~\cite{shen2025zoomeye,zhao2025pyvision}, with much less focus on tool-use efficiency. In practice, many open-source multimodal agents overuse tools whenever they are available, even when direct reasoning is sufficient. We term this failure mode \emph{blind tool-use reasoning}, and study how to train multimodal agents to use tools more selectively.

\begin{figure*}[!t]
	\centering
	\includegraphics[width=1\linewidth]{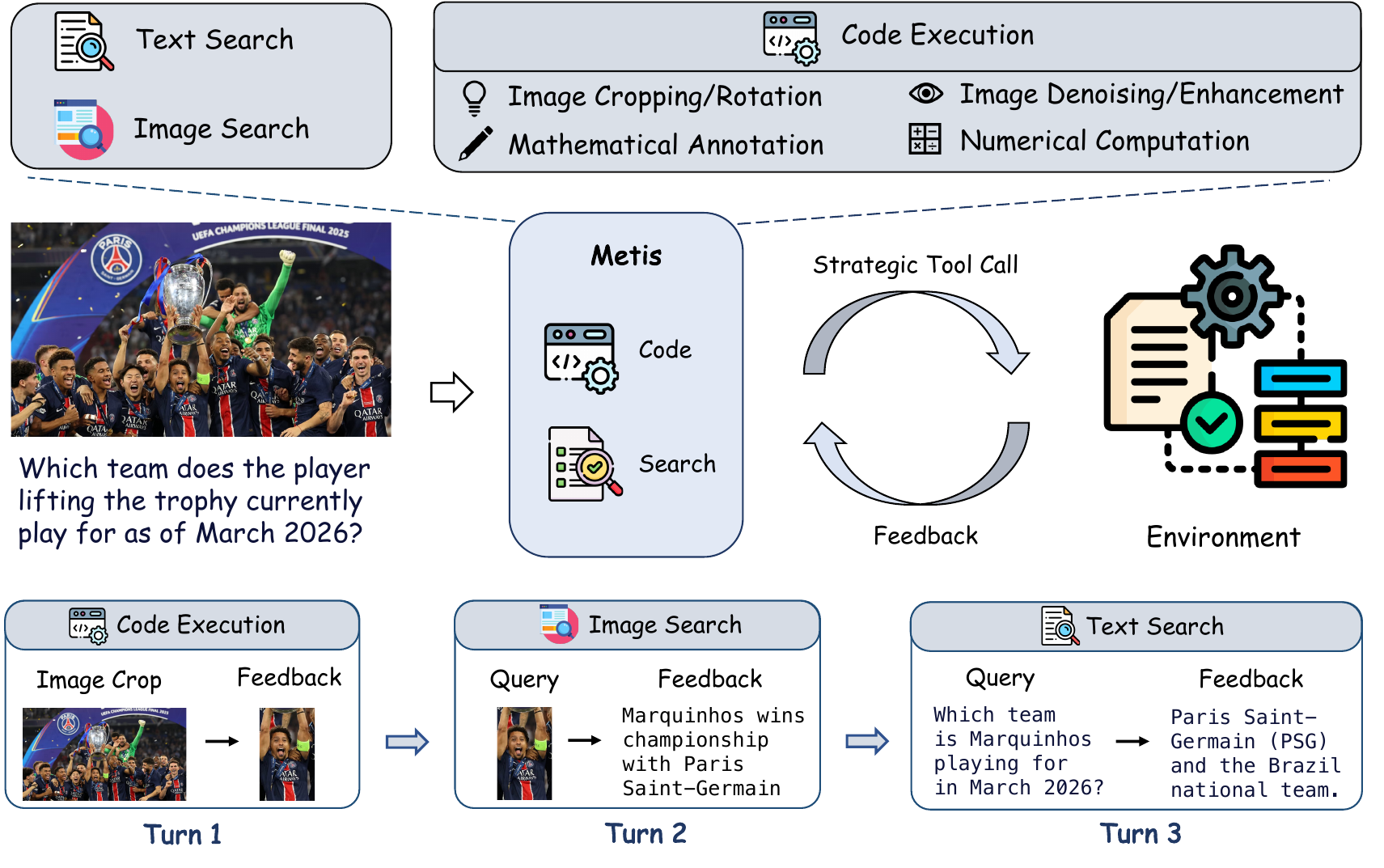}
	\caption{\textbf{Overview of \metis{}.} A strategic multimodal reasoning agent that selectively invokes code execution, text search, and image search tools during multi-turn reasoning. Rather than invoking tools by default, \metis{} adaptively determines when tool interactions provide genuinely useful evidence, and otherwise reasons directly from the available context to obtain the final answer.}
	\label{Fig.method} 
\end{figure*}

\section{Method}
\label{sec:method}

The overview of \metis{} is shown in Figure~\ref{Fig.method}. We begin by formalizing the multi-turn tool-augmented reasoning setting and analyzing the inherent flaws of existing coupled reward formulations (\S\ref{sec:problem_and_coupling}). We then present the Hierarchical Decoupled Policy Optimization (\hdpo{}) framework (\S\ref{sec:hdpo}), a method that eliminates cross-objective interference and naturally induces a learning curriculum.

% ========================================
% 3.1 Problem Formulation & The Coupling Problem
% ========================================
\subsection{Problem Formulation \& The Reward Coupling Problem}
\label{sec:problem_and_coupling}

Consider a multimodal language model with policy $\pi_\theta$ that answers visual reasoning queries by interleaving chain-of-thought reasoning with an external tool environment. Given a prompt, the model generates a group of $G$ multi-turn responses $\{y_1, y_2, \ldots, y_G\}$ (for simplicity, we omit the prompt index in this subsection and focus on a single group), where each response $y_i$ contains $T_i$ tool interactions before yielding a final answer. 

To jointly encourage accurate answers and efficient tool use, a straightforward application of GRPO~\cite{shao2024deepseekmath} would define a scalarized, coupled reward for each response $i$:
\begin{equation}
	R_i^{\mathrm{mix}} = R_i^{\mathrm{acc}} + \alpha \cdot R_i^{\mathrm{tool}}
	\label{eq:coupled_reward}
\end{equation}
where $R_i^{\mathrm{acc}}$ captures correctness and formatting, $R_i^{\mathrm{tool}}$ rewards tool parsimony, and $\alpha$ balances the two. This combined reward is then used to compute the advantage for policy optimization:
\begin{equation}
	A_i^{\mathrm{mix}} = \frac{R_i^{\mathrm{mix}} - \mathrm{mean}(\{R_1^{\mathrm{mix}}, \dots, R_G^{\mathrm{mix}}\})}{\mathrm{std}(\{R_1^{\mathrm{mix}}, \dots, R_G^{\mathrm{mix}}\})}
	\label{eq:coupled_advantage}
\end{equation}

While seemingly straightforward, this scalarization introduces a critical vulnerability: \emph{the shared advantage normalization entangles the two objectives, leading to severe credit misassignment}. Expanding the variance of the mixed reward via the linearity of variance yields:
\begin{equation}
	\mathrm{Var}(R^{\mathrm{mix}}) = \sigma_{\mathrm{acc}}^2 + \alpha^2 \sigma_{\mathrm{tool}}^2 + 2\alpha\, \mathrm{Cov}(R^{\mathrm{acc}}, R^{\mathrm{tool}})
	\label{eq:sigma_mix}
\end{equation}
where $\sigma_{\mathrm{acc}}^2$ and $\sigma_{\mathrm{tool}}^2$ denote the variances of the accuracy and tool rewards, respectively. Because correctness and tool use are inherently correlated, $\mathrm{Cov}(R^{\mathrm{acc}}, R^{\mathrm{tool}})$ is generally non-zero. As a result, the two objectives become mathematically entangled, precipitating three concrete pathologies:

\noindent \textbf{Gradient Entanglement:} The shared denominator inextricably links the policy gradients of the two objectives. The magnitude of the accuracy update becomes inversely proportional to the variance of the tool usage, and vice versa, causing destructive interference.

\noindent \textbf{Semantic Ambiguity:} A correct-but-inefficient trajectory may yield a scalar reward mathematically indistinguishable from an incorrect-but-efficient one. This conflation produces near-zero advantages for both, effectively neutralizing the training signal for critical edge cases.

\noindent \textbf{Hyperparameter Fragility:} The effective optimization trade-off is dictated not merely by $\alpha$, but by the highly dynamic, data-dependent covariance structure $\mathrm{Cov}(R^{\mathrm{acc}}, R^{\mathrm{tool}})$, rendering the hyperparameter notoriously unstable across diverse task distributions.

When the trade-off hyperparameter $\alpha$ is small, the efficiency signal is severely suppressed during advantage normalization. Specifically, let $\tilde{R}_i^{\mathrm{acc}}$ and $\tilde{R}_i^{\mathrm{tool}}$ denote the centered rewards (e.g., $\tilde{R}_i^{\mathrm{acc}} = R_i^{\mathrm{acc}} - \mathrm{mean}(R^{\mathrm{acc}})$). The mixed advantage expands to:
\begin{equation}
	A_i^{\mathrm{mix}} = \frac{\tilde{R}_i^{\mathrm{acc}} + \alpha \tilde{R}_i^{\mathrm{tool}}}{\sqrt{\sigma_{\mathrm{acc}}^2 + 2\alpha \,\mathrm{Cov}(R^{\mathrm{acc}},R^{\mathrm{tool}}) + \alpha^2 \sigma_{\mathrm{tool}}^2}}
	\label{eq:adv_mix_expanded}
\end{equation}
For sufficiently small $\alpha$, applying a first-order Taylor expansion reveals that the denominator is overwhelmingly dominated by the accuracy variance $\sigma_{\mathrm{acc}}$:
\begin{equation}
	A_i^{\mathrm{mix}} = \frac{\tilde{R}_i^{\mathrm{acc}}}{\sigma_{\mathrm{acc}}} + \mathcal{O}(\alpha)
	\label{eq:adv_taylor}
\end{equation}
This derivation explicitly reveals that the gradient contribution from tool efficiency is not only bounded by $\mathcal{O}(\alpha)$, but also heavily attenuated by the typically large accuracy variance $\sigma_{\mathrm{acc}}$. As $\alpha$ decreases to prevent accuracy degradation, the optimization signal for tool efficiency vanishes asymptotically. This mathematical reality explains why coupled-reward approaches fundamentally fail to curb blind tool invocation.

% ========================================
% 3.2 HDPO Framework
% ========================================
\subsection{\hdpo{}: Hierarchical Decoupled Policy Optimization}
\label{sec:hdpo}

\hdpo{} resolves the coupling problem by maintaining two \emph{independent} optimization channels. Instead of combining rewards before normalization, we compute separate advantages for accuracy and efficiency, each grounded in its own semantic baseline.

\subsubsection{Dual Reward Design and Decoupled Advantages}
\label{sec:dual_reward_adv}

We define two orthogonal rewards and compute their group-relative advantages independently.

\textbf{Accuracy Channel.} The accuracy reward $R_i^{\mathrm{acc}}$ evaluates the final response quality, comprising a correctness score and a format compliance bonus:
\begin{equation}
	R_i^{\mathrm{acc}} = \lambda_a \cdot R_i^{\mathrm{ans}} + \lambda_f \cdot R_i^{\mathrm{fmt}}
	\label{eq:r_acc}
\end{equation}
where $R_i^{\mathrm{ans}} \in \{0, 1\}$ is a binary correctness score from an LLM judge, $R_i^{\mathrm{fmt}} \in \{0, 1\}$ indicates format compliance, and we set $\lambda_a{=}0.9$, $\lambda_f{=}0.1$. To optimize this objective, we apply the standard GRPO advantage estimation over all $G$ rollouts in the group:
\begin{equation}
	A_i^{\mathrm{acc}} = \frac{R_i^{\mathrm{acc}} - \mathrm{mean}(\{R_1^{\mathrm{acc}}, \dots, R_G^{\mathrm{acc}}\})}{\mathrm{std}(\{R_1^{\mathrm{acc}}, \dots, R_G^{\mathrm{acc}}\}) + \epsilon}
	\label{eq:a_acc}
\end{equation}
where $\epsilon$ is a small constant to ensure numerical stability.

\textbf{Efficiency Channel.} To counteract latency-agnostic behavior, the tool reward explicitly optimizes for execution economy (i.e., tool parsimony). However, to prevent the agent from gaming the reward function by prematurely terminating trajectories, this efficiency signal must be strictly \emph{conditioned on correctness}. An incorrect rollout must never be rewarded for mere alacrity. Thus, we define:
\begin{equation}
	R_i^{\mathrm{tool}} = 
	\begin{cases}
		\displaystyle \frac{1}{T_i + 1} & \text{if } R_i^{\mathrm{ans}} > 0, \\[6pt]
		0 & \text{otherwise}.
	\end{cases}
	\label{eq:r_tool}
\end{equation}
where $T_i$ denotes the number of tool invocations in the $i$-th rollout. This inverse penalty yields a monotonically decreasing reward as the number of tool calls increases ($T{=}0 \mapsto 1.0$, $T{=}1 \mapsto 0.5$, etc.), heavily penalizing redundant interactions while preserving a smooth preference structure.

However, na\"ively applying standard GRPO over all $G$ rollouts for $R_i^{\mathrm{tool}}$ would pull the group mean toward zero due to the presence of incorrect rollouts (which are assigned $R_i^{\mathrm{tool}}=0$). This artificially inflates the advantage of any correct rollout, regardless of its actual efficiency. To circumvent this, we employ a \textbf{conditional advantage estimation} mechanism. We define a qualifying set $\mathcal{Q}$ of indices corresponding exclusively to correct responses:
\begin{equation}
	\mathcal{Q} = \{j \in \{1 \dots G\} \mid R_j^{\mathrm{ans}} > 0\}
	\label{eq:qualifying_set}
\end{equation}
The tool efficiency advantage is then computed \emph{exclusively} relative to other correct solutions:
\begin{equation}
	A_i^{\mathrm{tool}} = 
	\begin{cases}
		\displaystyle \frac{R_i^{\mathrm{tool}} - \mathrm{mean}(\{R_k^{\mathrm{tool}}\}_{k \in \mathcal{Q}})}{\mathrm{std}(\{R_k^{\mathrm{tool}}\}_{k \in \mathcal{Q}}) + \epsilon} & \text{if } i \in \mathcal{Q} \text{ and } |\mathcal{Q}| \geq 2, \\[8pt]
		0 & \text{otherwise}.
	\end{cases}
	\label{eq:a_tool}
\end{equation}
When fewer than two rollouts are correct ($|\mathcal{Q}| < 2$), no meaningful within-group comparison of tool efficiency exists. In such cases, we assign zero advantage to prevent semantically invalid cross-prompt comparisons, thereby ensuring that the efficiency signal remains strictly grounded in intra-task relative performance.

\subsubsection{Hierarchical Policy Update}
\label{sec:dual_loss}

With the advantages cleanly decoupled, we construct the final \hdpo{} objective by linearly combining their respective clipped surrogate losses. Let $\mathcal{L}_{\mathrm{GRPO}}(A)$ denote the standard PPO-style clipped surrogate objective~\cite{schulman2017proximal} for a given advantage $A$. The joint policy gradient loss is formulated as:
\begin{equation}
	\mathcal{L}_{\mathrm{HDPO}}(\theta) = w_{\mathrm{acc}} \cdot \mathcal{L}_{\mathrm{GRPO}}\!\left(A^{\mathrm{acc}}\right) + w_{\mathrm{tool}} \cdot \mathcal{L}_{\mathrm{GRPO}}\!\left(A^{\mathrm{tool}}\right)
	\label{eq:loss_hdpo}
\end{equation}
Because $A^{\mathrm{acc}}$ and $A^{\mathrm{tool}}$ are normalized independently across distinct semantic baselines, the policy gradient decomposes cleanly. Each gradient component delivers a targeted, orthogonal learning signal for its respective objective, entirely eliminating the destructive covariance interference observed in Eq.~\ref{eq:sigma_mix}. Crucially, this orthogonalization allows us to impose a meaningful efficiency penalty ($w_{\mathrm{tool}}$) without risking the catastrophic degradation of task accuracy that plagues coupled formulations.

\subsubsection{Algorithm Summary \& The Implicit Curriculum}
\label{sec:algorithm_summary}

Algorithm~\ref{alg:hdpo} summarizes the complete \hdpo{} procedure. In each iteration, the policy samples multiple rollouts per prompt through interaction with the tool environment. It computes two orthogonal rewards: an accuracy reward for task correctness and a tool reward for execution efficiency. Next, it estimates $\hat{A}_{\mathrm{acc}}$ using standard GRPO over all rollouts in each prompt group, while estimating $\hat{A}_{\mathrm{tool}}$ exclusively over the qualifying set of correct rollouts. The policy is finally updated via the weighted sum of the two surrogate losses. 

A notable emergent property of this decoupled, conditional design is the induction of an \emph{implicit cognitive curriculum}. Early in training, when the policy struggles with the task, the qualifying set $\mathcal{Q}$ is predominantly empty. Consequently, the optimization is naturally dominated by the accuracy objective, forcing the model to prioritize functional correctness. As the model's reasoning capabilities mature, more rollouts qualify for the efficiency comparison ($|\mathcal{Q}| \geq 2$), smoothly scaling up the tool-parsimony signal. This dynamic elegantly enforces a two-phase developmental trajectory---\emph{first learn to be correct, then learn to be efficient}---without necessitating any explicit, manual reward scheduling or hyperparameter annealing.

% ========================================
% Algorithm
% ========================================

\begin{algorithm}[t]
	\caption{\hdpo{}: Hierarchical Decoupled Policy Optimization}
	\label{alg:hdpo}
	\begin{algorithmic}[1]
		% 定义输入输出的显示方式
		\renewcommand{\algorithmicrequire}{\textbf{Input:}}
		\renewcommand{\algorithmicensure}{\textbf{Output:}}
		
		\Require Policy $\pi_\theta$, prompt set $\{x_i\}$, rollouts per prompt $G$, weights $w_{\mathrm{acc}}$ and $w_{\mathrm{tool}}$, environment $\mathcal{E}$
		
		\For{each training iteration}
		\State \textbf{Rollout:} For each $x_i$, sample $G$ trajectories $\{y_i^{(j)}\}_{j=1}^G$ via multi-turn interaction with $\mathcal{E}$
		\State \textbf{Reward:} Compute $R_{\mathrm{acc}}^{(i,j)}$ (Eq.~\ref{eq:r_acc}) and $R_{\mathrm{tool}}^{(i,j)}$ (Eq.~\ref{eq:r_tool}) for each rollout
		\State \textbf{Accuracy advantage:} $\hat{A}_{\mathrm{acc}}^{(i,j)} \gets \text{GRPO}(\{R_{\mathrm{acc}}^{(i,k)}\}_{k=1}^G)$ over all $G$ rollouts per group \hfill $\triangleright$ Eq.~\ref{eq:a_acc}
		\State \textbf{Qualifying set:} $\mathcal{Q}_i \gets \{j : R_{\mathrm{ans}}^{(i,j)} > 0\}$ for each prompt $x_i$ \hfill $\triangleright$ Eq.~\ref{eq:qualifying_set}
		\State \textbf{Tool advantage:} Compute $\hat{A}_{\mathrm{tool}}^{(i,j)}$ using Eq.~\ref{eq:a_tool} over the qualifying set $\mathcal{Q}_i$
		\State \textbf{Update:} $\theta \gets \theta - \eta\, \nabla_\theta \left[w_{\mathrm{acc}} \cdot \mathcal{L}_{\mathrm{GRPO}}(\hat{A}_{\mathrm{acc}}) + w_{\mathrm{tool}} \cdot \mathcal{L}_{\mathrm{GRPO}}(\hat{A}_{\mathrm{tool}})\right]$ \hfill $\triangleright$ Eq.~\ref{eq:loss_hdpo}
		\EndFor
	\end{algorithmic}
\end{algorithm}

\subsection{Training Data Curation}
\label{sec:data}

A mathematically rigorous RL framework requires an equally robust empirical foundation. While \hdpo{} resolves the credit assignment problem during optimization, the policy's ultimate behavior is bottlenecked by the semantic integrity of the behavioral priors (SFT) and the validity of the environmental feedback (RL). We identify pervasive pathologies in existing tool-augmented MLLM datasets---specifically, hallucinated environmental dynamics and obsolete tool dependencies---and propose a rigorous, meta-cognitive curation pipeline.

\subsubsection{SFT Data Curation}
\label{sec:sft_data}

Our SFT corpus is sourced from publicly available tool-augmented multimodal trajectories~\cite{hong2025deepeyesv2,qiao2025v,zhang2025openmmreasoner,zhang2025thyme}. We identify and eradicate low-quality samples through three targeted mechanisms:

\noindent\textbf{Eradicating Hallucinated Environmental Dynamics.} A pervasive flaw in existing SFT demonstrations is the presence of non-executable code (e.g., syntax errors, missing dependencies) coupled with hallucinated tool observations. In such corrupted trajectories, the environment either miraculously returns a correct output for broken code, or the agent blatantly ignores a runtime error and hallucinates the correct final answer. Training on these trajectories severely damages the model's grounding, teaching it to exploit environmental loopholes rather than perform genuine reasoning. To rectify this, we rigorously execute all code segments within a sandboxed environment, strictly discarding any trajectory that exhibits execution failures or feedback inconsistencies.

\noindent\textbf{Isolating Genuine Tool Necessity.} Many existing datasets were annotated using weaker baseline models that relied on external tools for relatively simple queries. As intrinsic model capabilities (e.g., internal parametric knowledge) advance, retaining these legacy annotations actively conditions the model to exhibit blind tool invocation. To enforce tool parsimony, we establish a zero-shot solvability baseline by evaluating the base model (Qwen3-VL-8B~\cite{bai2025qwen3}) on candidate samples using direct reasoning (without tool access). Samples that are consistently solved correctly (pass@8 = 1) are aggressively filtered out, ensuring the SFT phase only demonstrates tool usage when strictly necessary.

\noindent\textbf{Multidimensional Meta-Cognitive Filtering.} Beyond mere execution correctness, the semantic quality of the reasoning chain is paramount. We employ Gemini-3.1-Pro~\citep{gemini3_1} as an automated judge to evaluate trajectories across multiple fine-grained dimensions (e.g., visual relevance, reasoning coherence, and tool-use rationale). Crucially, the judge explicitly penalizes ``blind tool invocation''---such as applying meaningless image rotations to an already clear image. Trajectories failing to meet stringent quality thresholds are discarded, ensuring the SFT corpus exclusively exemplifies strategic, meta-cognitive tool use.

\subsubsection{RL Data Curation}
\label{sec:rl_data}

For the RL stage, we curate a prompt set from multiple datasets~\cite{chng2025sensenova,hong2025deepeyesv2,qiao2025v,zhang2025thyme}, covering diverse task types including mathematical reasoning, fine-grained visual understanding, and search-oriented tasks. We apply the following filtering criteria to guarantee a high-fidelity reward signal:

\noindent\textbf{Environmental Fidelity Verification.} To ensure the RL environment provides a stable and meaningful optimization signal, we pass raw prompts through the multimodal judge to assess image quality, question clarity, and image-text consistency. Prompts with corrupted visual inputs or severe semantic ambiguity are excluded, preventing the policy from fitting to noise.

\noindent\textbf{Variance-Aware Difficulty Calibration.} Prompts that are trivially easy (solved by all $G$ rollouts) or prohibitively hard (solved by none) yield zero-variance accuracy rewards, leading to degenerate advantage estimates in GRPO. We empirically sample $G=8$ rollouts per prompt and strictly retain only those exhibiting a non-trivial mix of successes and failures, guaranteeing a robust and actionable gradient signal for the policy update.

\section{Experiments}

\subsection{Experimental Setup}

\paragraph{Training Datasets.}
Our SFT corpus is sourced from publicly available tool-augmented multimodal trajectories, including DeepEyesV2~\citep{hong2025deepeyesv2}, V-Interaction~\citep{qiao2025v}, and Thyme~\citep{zhang2025thyme}. We rigorously apply the three-stage meta-cognitive curation pipeline detailed in \S\ref{sec:sft_data}: (i) eradicating hallucinated environmental dynamics, (ii) isolating genuine tool necessity by filtering out samples where the base model achieves $\text{pass@8}=1$ under direct reasoning, and (iii) applying multidimensional meta-cognitive filtering. To preserve intrinsic reasoning capabilities, we additionally incorporate tool-free reasoning data from OpenMMReasoner~\citep{zhang2025openmmreasoner}. For the RL stage, we curate a prompt set from V-Interaction~\citep{qiao2025v}, Thyme~\citep{zhang2025thyme}, SenseNova-MARS~\citep{chng2025sensenova}, and DeepEyesV2~\citep{hong2025deepeyesv2}. We strictly retain only samples with $\text{pass@8} \in (0, 1)$ to ensure a non-trivial, variance-aware training signal. The final RL training set comprises about 5K high-quality prompts covering diverse task types: perception-related data (45\%), search-oriented data (36\%), and mathematical/general reasoning tasks (19\%).

\paragraph{Implementation Details.}
We train \metis{} using Qwen3-VL-8B-Instruct~\citep{bai2025qwen3} as the backbone model. The training proceeds in two stages: supervised fine-tuning (SFT) for cold-start initialization, followed by reinforcement learning (RL) via \hdpo{}. During SFT, we train for 2 epochs using the AdamW optimizer with a cosine learning rate decay, a peak learning rate of $1\times10^{-5}$, and a global batch size of 128. During the RL stage, we optimize the policy using \hdpo{} with a batch size of 128, sampling $G{=}16$ rollouts per prompt. The learning rate is set to $1\times10^{-6}$, and the KL penalty coefficient is strictly set to 0 to encourage extensive exploration. The maximum response length is capped at 16,384 tokens to accommodate complex, multi-turn tool interactions. For the dual-channel optimization, we set the loss weights to $w_{\mathrm{acc}}{=}1.0$ and $w_{\mathrm{tool}}{=}0.15$. And all experiments were performed on a server featuring 8 NVIDIA Blackwell B200 GPUs.

\paragraph{Baselines.}
We compare \metis{} against three categories of strong baselines: (1) \textit{Open-source models without tool use}, including LLaVA-OneVision~\citep{li2024llava}, InternVL3-8B~\citep{zhu2025internvl3}, Qwen2.5-VL-7B/32B-Instruct~\citep{bai2025qwen2}, and Qwen3-VL-8B-Instruct~\citep{bai2025qwen3}; (2) \textit{Text-only reasoning models}, including MM-Eureka~\citep{meng2025mm}, ThinkLite-VL~\citep{wang2025sota}, VL-Rethinker~\citep{wang2025vl}, and VLAA-Thinker~\citep{chen2025sft}; and (3) \textit{Agentic multimodal models}, including Pixel-Reasoner~\citep{wang2025pixel}, DeepEyes~\citep{zheng2025deepeyes}, Thyme~\citep{zhang2025thyme}, DeepEyesV2~\citep{hong2025deepeyesv2}, Mini-o3~\citep{lai2025mini}, and Skywork-R1V4-30B-A3B~\citep{zhang2025skywork}.

\paragraph{Benchmarks.}
We evaluate \metis{} across two broad groups of benchmarks covering complementary cognitive capabilities. \textbf{Perception and Document Understanding:} V*Bench~\citep{wu2024v}, HRBench-4K/8K~\citep{wang2025divide}, TreeBench~\citep{}, MME-RealWorld~\citep{zhang2024mme}, SEEDBench2-Plus~\citep{li2024seed}, and CharXiv (descriptive and reasoning questions)~\citep{wang2024charxiv}. \textbf{Mathematical and Logical Reasoning:} MathVista$_{\text{mini}}$~\citep{lu2023mathvista}, MathVerse$_{\text{mini}}$~\citep{zhang2024mathverse}, WeMath~\citep{qiao2025we}, DynaMath~\citep{zou2024dynamath}, and LogicVista~\citep{xiao2024logicvista}.

\begin{table*}[t]
	\centering
	\setstretch{1.3}
    \caption{\textbf{Performance comparison on visual perception and document understanding benchmarks.} \metis{} consistently outperforms existing open-source agentic models, demonstrating that strategic tool use enhances performance on high-resolution and complex document tasks.}
	\resizebox{\linewidth}{!}{
		\begin{tabular}{l | c c c c c | c c c}
			\toprule
			\multirow{2}{*}{\textbf{Models}} & \multicolumn{5}{c}{\textbf{Perception}} & \multicolumn{3}{|c}{\textbf{Document}} \\
			\cline{2-9}
			& V* Bench & HR4K & HR8K &TreeBench & MME RealWorld & SEED2 PLUS &CharXiv(DQ)  &CharXiv(RQ) \\
			\midrule
			
			\multicolumn{9}{c}{\textit{\textbf{Open-Source Models}}} \\
			\midrule
			LLaVA-OneVision~\citep{li2024llava} & 75.4 & 63.0 & 59.8 &37.3 & 57.4 & 65.4 &- & - \\
			InternVL3-8B~\citep{zhu2025internvl3} & 81.2 & 70.0 & 69.3 &38.8 & - & 69.7 &73.6 & 37.6 \\
			Qwen2.5-VL-7B-Instruct~\citep{bai2025qwen2} & 75.3 & 65.5 & 62.1 &37.0 & 56.8 & 70.4 &72.7 & 40.2 \\
			Qwen2.5-VL-32B-Instruct~\citep{bai2025qwen2} & 80.6 & 69.3 & 63.6 &42.5& 59.1 & 72.4 &83.2 & 48.0 \\
			Qwen3-VL-8B-Instruct~\citep{bai2025qwen3} & 86.4 & 78.9 & 74.6 &40.7 & 61.9 &71.0 &83.0 &46.3 \\
			\midrule
			
			\multicolumn{9}{c}{\textit{\textbf{Agentic Multimodal Models}}} \\
			\midrule
			Pixel-Reasoner~\citep{wang2025pixel} & 84.3 & 72.6 & 66.1 &39.0 & 64.4 & - & - & -\\
			DeepEyes~\citep{zheng2025deepeyes} & 83.3 & 73.2 & 69.5 &37.5 & 64.1 & - & - & -\\
			Thyme~\citep{zhang2025thyme} & 82.2 & 77.0 & 72.0 &- & 64.8 & - & - & -\\
			DeepEyesV2~\citep{hong2025deepeyesv2} & 81.8 & 77.9 & 73.8 &42.5 & 64.9 & 70.5 &78.6 & 48.9 \\
			Mini o3~\citep{lai2025mini} & 88.2 & 77.5 & 73.3 &-& 65.5 & - & - & -\\
			SenseNova-MARS-8B~\citep{chng2025sensenova} &\textbf{92.2}&83.1&78.4&-&67.9&-&-& -\\
			Skywork-R1V4-30B-A3B~\citep{zhang2025skywork} & 88.0 & 82.8 & 79.8 &-& \textbf{71.4} & - & - & -\\
			\midrule
			
			\rowcolor{blue!8}
			\textbf{\metis{}} & 91.1 & \textbf{83.5} & \textbf{82.0} &\textbf{45.2} & 70.3& \textbf{72.5} &\textbf{83.4} & \textbf{54.1} \\
			\bottomrule
		\end{tabular}
	}
	\label{tab:percept}
\end{table*}

\subsection{Main Results}

We present a comprehensive evaluation of \metis{} across perception, document understanding, and mathematical reasoning benchmarks. As shown in Table~\ref{tab:percept} and Table~\ref{tab:math_reasoning}, \metis{} establishes new state-of-the-art or highly competitive performance across a wide range of metrics among open-source multimodal agents, demonstrating that strategic tool use directly translates to superior reasoning outcomes.

\paragraph{Perception and Document Understanding.} 
Table~\ref{tab:percept} details the performance on tasks requiring fine-grained visual inspection and document parsing. \metis{} achieves remarkable improvements over its strong backbone, Qwen3-VL-8B-Instruct. Notably, on high-resolution benchmarks like HRBench-4K and HRBench-8K, \metis{} attains 83.5\% and 82.0\% respectively, outperforming all existing agentic models including the 30B-parameter Skywork-R1V4. Furthermore, on the highly challenging CharXiv reasoning questions, \metis{} achieves 54.1\%, significantly surpassing the previous best agentic model (DeepEyesV2 at 48.9\%). This demonstrates that our meta-cognitive training enables the agent to effectively leverage image cropping and search tools to resolve visual ambiguities that stump standard models.

\paragraph{Mathematical and Logical Reasoning.}
Table~\ref{tab:math_reasoning} highlights \metis{}'s capabilities on rigorous mathematical and logical reasoning benchmarks. \metis{} achieves an outstanding average score of 66.9\% across five demanding datasets, substantially outperforming both text-only reasoning models and existing agentic multimodal models. Particularly striking is the performance on WeMath (65.2\%), where \metis{} achieves a massive absolute improvement of +26.4\% over its backbone (38.8\%) and eclipses previous agents like DeepEyesV2 (38.1\%). This substantial leap underscores the efficacy of \hdpo{}: by eliminating gradient entanglement, the model learns to seamlessly interleave Python code execution for complex calculations without compromising its core logical reasoning chain.

\begin{table*}[t]
	\centering
	\setstretch{1.1}
    \caption{\textbf{Performance comparison on multimodal reasoning benchmarks.} By decoupling the efficiency penalty, \metis{} effectively leverages code execution for complex calculations, achieving state-of-the-art accuracy across all evaluated datasets.}
	\resizebox{\linewidth}{!}{
		\begin{tabular}{l | c c c c c | c}
			\toprule
			\textbf{Models} & MathVista$_{\text{mini}}$ & MathVerse$_{\text{mini}}$ & WeMath & DynaMath & LogicVista & Avg. \\
			\midrule
			
			\multicolumn{7}{c}{\textit{\textbf{Open-source Models}}} \\
			\midrule
			LLaVA-OneVision~\citep{li2024llava} & 58.6 & 19.3 & 20.9 & -   & 33.3 & - \\
			Qwen-2.5-VL-7B-Instruct~\citep{bai2025qwen2} & 68.3 & 45.6 & 34.6 & 53.3 & 45.9 & 49.5 \\
			InternVL3-8B~\citep{zhu2025internvl3}  & 71.6 & 39.8 & 37.1 & -   & 44.1 & - \\
			Qwen3-VL-8B-Instruct~\citep{bai2025qwen3} & 76.3 & 61.3 & 38.8 & 65.5 & 54.9 & 59.4 \\
			
			\midrule
			\multicolumn{7}{c}{\textit{\textbf{Text-only Reasoning Models}}} \\
			\midrule
			MM-Eureka-7B~\citep{meng2025mm} & 72.6 & 50.3 & 21.8 & -   & 46.3 & - \\
			ThinkLite-VL-7B~\citep{wang2025sota} & 75.1 & 52.1 & 41.8 & -   & 42.7 & - \\
			VL-Rethinker-7B~\citep{wang2025vl} & 74.9 & 54.2 & 36.3 & -   & 42.7 & - \\
			VLAA-Thinker-7B~\citep{chen2025sft} & 71.7 & -   & 35.7 & -   & 45.9 & - \\
			
			\midrule
			\multicolumn{7}{c}{\textit{\textbf{Agentic Multimodal Models}}} \\
			\midrule
			DeepEyes~\citep{zheng2025deepeyes} & 70.1 & 47.3 & 38.9 & 55.0 & 47.7 & 51.8 \\
			Thyme~\citep{zhang2025thyme} & 70.0 & -   & 39.3 & -   & 49.0 & - \\
			DeepEyesV2~\citep{hong2025deepeyesv2} & 71.9 & 52.7 & 38.1 & 57.2 & 48.7 & 53.7 \\
			
			\midrule
			\rowcolor{blue!8}
			\textbf{\metis{} } & \textbf{78.0} & \textbf{65.9} & \textbf{65.2} & \textbf{69.2} & \textbf{56.2} & \textbf{66.9} \\
			\bottomrule
		\end{tabular}
	}
	\label{tab:math_reasoning}
\end{table*}

\subsection{Ablation Studies}

To systematically validate the contributions of our framework, we conduct ablation studies on a representative subset of benchmarks. Table~\ref{tab:ablation_hdpo} reports the performance under identical backbone and training data configurations. Note that setting $w_{\mathrm{tool}}{=}0$ gracefully degrades \hdpo{} to standard GRPO, where only the accuracy objective is optimized.

\paragraph{Effectiveness of Decoupled Optimization.}
Compared to the base model (Qwen3-VL-8B-Instruct), standard GRPO ($w_{\mathrm{tool}}{=}0$) yields noticeable improvements across all benchmarks, confirming the general benefits of RL fine-tuning. However, the introduction of our decoupled tool-efficiency objective (\hdpo{}) unlocks substantially higher performance ceilings. Specifically, \hdpo{} ($w_{\mathrm{tool}}{=}0.15$) achieves absolute gains of +2.4\%, +2.8\%, and +3.1\% over standard GRPO on V* Bench, HRBench8K, and CharXiv (RQ), respectively. These results empirically validate our core hypothesis: task accuracy and tool efficiency are not inherently conflicting. By decoupling the two objectives and eliminating gradient entanglement, \hdpo{} successfully suppresses noisy, redundant tool invocations, which in turn consistently elevates the final reasoning accuracy.

\begin{table}[ht]
	\centering
	\setstretch{1.25}
    \caption{\textbf{Ablation study on the optimization objective and efficiency weight.} Setting $w_{\mathrm{tool}}{=}0$ reduces the optimization to standard GRPO (accuracy-only). \hdpo{} consistently improves reasoning performance across all evaluated benchmarks, demonstrating that strategic tool parsimony acts as a catalyst for accuracy.}
	\resizebox{\linewidth}{!}{
		\begin{tabular}{l | c c c c c}
			\toprule
			\textbf{Method} & \textbf{V* Bench} & \textbf{HRBench4K} & \textbf{HRBench8K} & \textbf{CharXiv(RQ)} & \textbf{MathVista$_{\text{mini}}$} \\
			\midrule
			Qwen3-VL-8B-Instruct &86.4 &78.9 &74.6 &46.3 &76.3\\
			+ standard GRPO ($w_{\mathrm{tool}}{=}0$) & 88.7 & 81.0 & 79.2 & 51.0 & 76.9 \\
			+ \hdpo{} ($w_{\mathrm{tool}}{=}0.10$) &88.0 & \textbf{83.5} &81.0 &52.7 & 77.4 \\
			\rowcolor{blue!8}
			+ \hdpo{} ($w_{\mathrm{tool}}{=}0.15$) & \textbf{91.1} & \textbf{83.5} & \textbf{82.0} & \textbf{54.1} & \textbf{78.0} \\
			+ \hdpo{} ($w_{\mathrm{tool}}{=}0.20$) &87.4 &82.5 &80.5 &51.5 & 77.2 \\
			\bottomrule
		\end{tabular}
	}
	\label{tab:ablation_hdpo}
\end{table}

\begin{figure*}[t]
	\centering
	\includegraphics[width=1\linewidth]{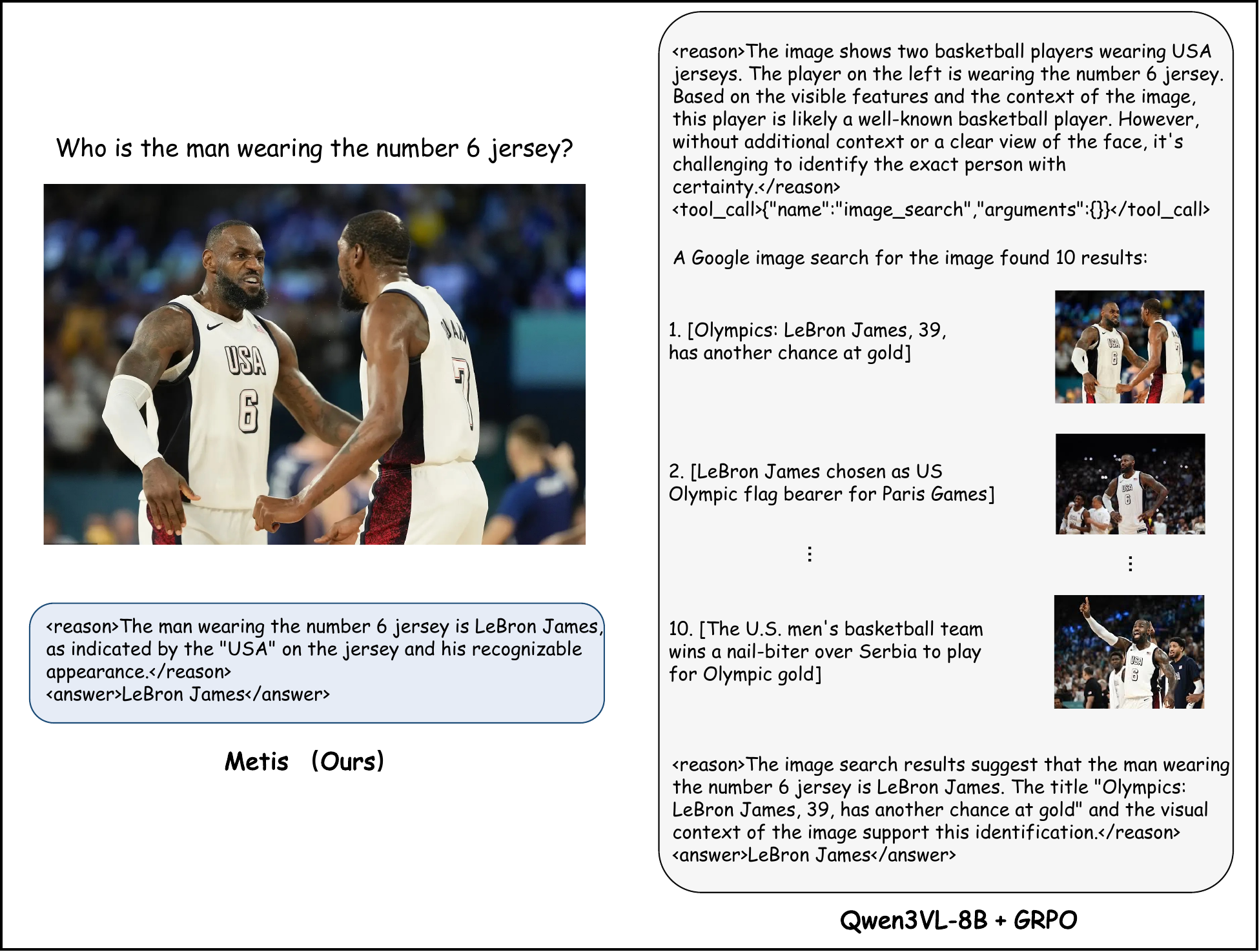}
	\caption{\textbf{Direct reasoning from visual context.} The query can be resolved through visual understanding and prior knowledge alone. \metis{} abstains from tool invocation and answers directly, exemplifying the meta-cognitive restraint instilled by \hdpo{}.}
	\label{Fig.main_case1} 
\end{figure*} 

\paragraph{Sensitivity to Efficiency Loss Weight.}
We further investigate the impact of the tool-efficiency weight $w_{\mathrm{tool}}$. A conservative weight ($w_{\mathrm{tool}}{=}0.10$) provides clear improvements over standard GRPO but remains suboptimal compared to $w_{\mathrm{tool}}{=}0.15$, suggesting that the efficiency signal is too weak to fully curb blind tool invocation. Conversely, an overly aggressive weight ($w_{\mathrm{tool}}{=}0.20$) precipitates performance degradation across all benchmarks. This indicates that overemphasizing tool parsimony forces the policy into a conservative regime, stifling necessary exploration and tool usage on arduous tasks. Overall, the performance exhibits a clear inverted-U trajectory, with $w_{\mathrm{tool}}{=}0.15$ striking the optimal meta-cognitive balance between self-reliance and external tool querying.

\begin{figure*}[t]
	\centering
	\includegraphics[width=1\linewidth]{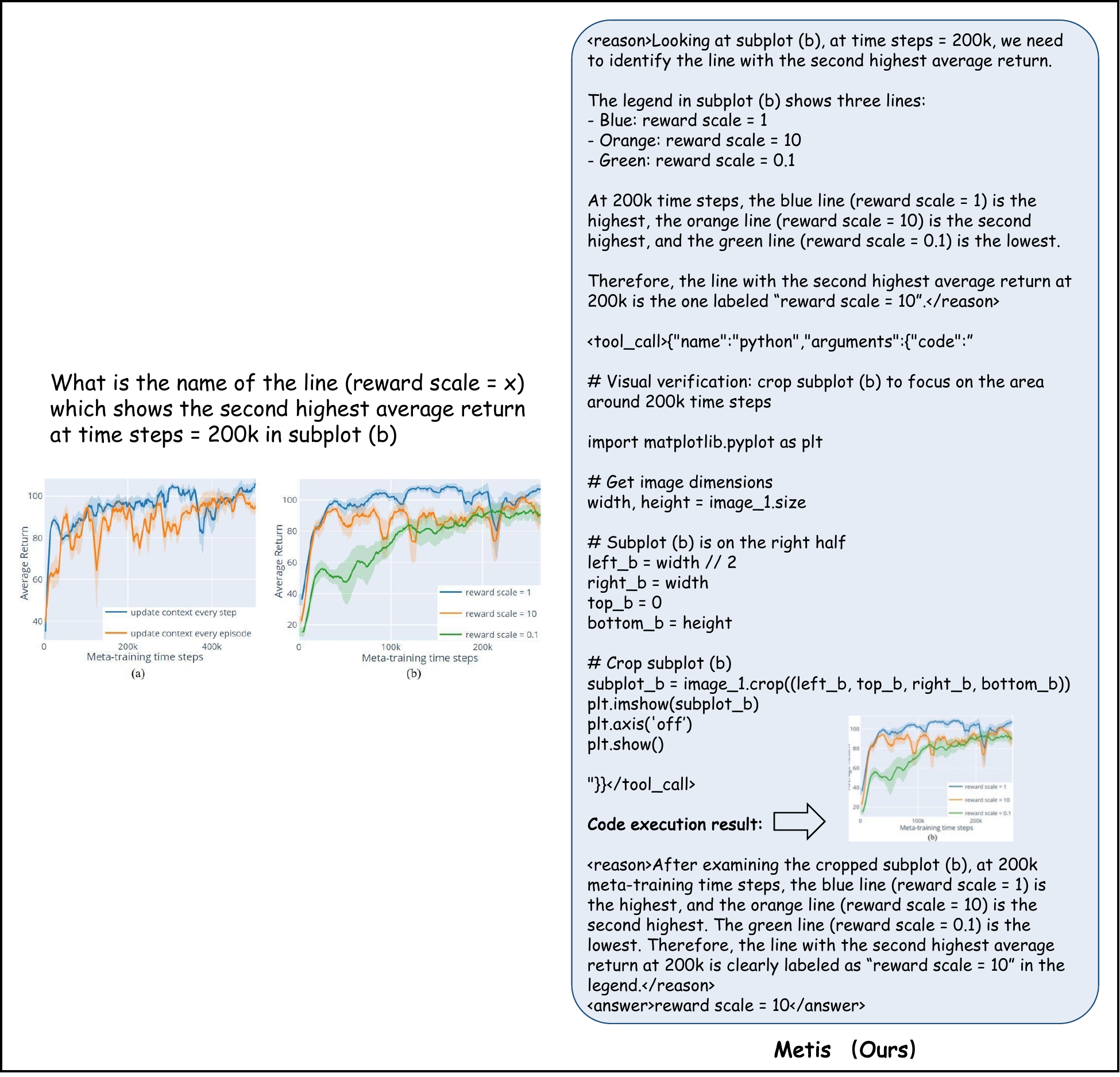}
	\caption{\textbf{Targeted code execution for fine-grained visual analysis.} The question requires comparing curves in a specific subplot region that is difficult to resolve at the original image scale. \metis{} invokes code to crop and enlarge the relevant area, enabling precise identification of the curve behavior near the queried time step.}
	\label{Fig.main_case2} 
\end{figure*} 

\subsection{Meta-Cognitive Tool Arbitration}
\label{sec:qualitative}

To complement the quantitative evaluation, we present representative cases that illustrate the meta-cognitive tool-use behavior cultivated by \hdpo{}. Figure~\ref{Fig.main_case1} shows a scenario where \metis{} resolves the query entirely through internal visual understanding and parametric knowledge, without resorting to any external tool. The agent directly infers the answer from the image content, exemplifying the core benefit of \hdpo{}: by penalizing unnecessary tool invocations within the efficiency channel, the agent learns to trust its own capabilities for queries within its competence, thereby avoiding the latency overhead and noise injection of redundant tool calls.

In contrast, Figure~\ref{Fig.main_case2} presents a scenario where fine-grained visual analysis exceeds the model's native resolution capabilities. Rather than guessing from the full image, \metis{} strategically invokes code execution to crop and enlarge the relevant subplot region, enabling precise inspection of overlapping curves and legend entries. This case highlights that \metis{} treats code execution not as a default fallback, but as a precision instrument deployed only when the visual evidence at the original resolution is genuinely ambiguous. Together, these two cases demonstrate that \metis{} has internalized a principled decision boundary: abstaining when internal knowledge suffices, and selectively engaging external tools only when genuinely necessary. Additional cases covering selective search tool invocation are provided in Appendix~\ref{sec:appendix_cases}.

\section{Conclusion}

In this work, we identify \textit{blind tool invocation} as a critical failure mode in tool-augmented MLLMs and propose Hierarchical Decoupled Policy Optimization (\hdpo{}) to address this meta-cognitive deficit. By decoupling task accuracy and tool efficiency into orthogonal channels via a conditional advantage mechanism, \hdpo{} eliminates gradient entanglement and naturally induces a cognitive curriculum. Complemented by a rigorous data curation pipeline, our resulting agent, \metis{}, reduces tool invocations by orders of magnitude while achieving state-of-the-art reasoning performance. Future work will explore extending this meta-cognitive framework to more open-ended, long-horizon environments. Ultimately, \metis{} challenges the paradigm of latency-agnostic scaling, proving that true intelligence lies not merely in knowing \textit{how} to interact with the world, but in possessing the meta-cognitive wisdom of \textit{when} to abstain.

\clearpage

{
	\small
	\bibliographystyle{plain}
	\bibliography{ref}
}
%\medskip
%
%

%%%%%%%%%%%%%%%%%%%%%%%%%%%%%%%%%%%%%%%%%%%%%%%%%%%%%%%%%%%%
\clearpage

%%%%%%%%%%%%%%%%%%%%%%%%%%%%%%%%%%%%%%%%%%%%%%%%%%%%%%%%%%%%
 \appendix
\section*{Appendix}

\section{System Prompt}
 
 The system prompt is presented in Figure~\ref{fig:system_prompt}. It explicitly defines the available tool, its calling format, and the execution environment, so that the model clearly understands how and when external code execution can be used. In addition, the prompt provides decision guidelines that encourage the model to reason before acting, answer directly whenever possible, and call the tool only when it is genuinely necessary. The required output structure is also specified through dedicated \texttt{<reason>}, \texttt{<tool\_call>}, and \texttt{<answer>} fields, which helps maintain consistent behavior and promotes efficient tool use.

\begin{figure*}[!h]
	\centering
	\begin{tcolorbox}[
		boxrule=1pt,
		boxsep=2pt,
		colback=gray!20,
		fontupper=\scriptsize,
		fonttitle=\scriptsize\bfseries,
		title=System Prompt
		]
		\setlist[itemize]{leftmargin=*, nosep, after=\vspace{4pt}}
		
		You are an efficient problem-solving agent. Your goal is to answer the user's question accurately while minimizing unnecessary tool usage.\\
		
		\textbf{\# Tools}\\
		You have access to the following tools. Use them ONLY when they provide clear value that reasoning alone cannot.\\
		
		\textbf{\#\# Python Code Execution}\\
		Write Python code to perform numerical analysis, data processing, or image operations (e.g., cropping, resizing, rotating, color adjustment, contrast enhancement, drawing auxiliary lines).
		
		\vspace{2pt}
		\textbf{Format:}\\
		\texttt{<tool\_call>\{"name": "python", "arguments": \{"code": "your code here"\}\}</tool\_call>}
		
		\vspace{2pt}
		\textbf{Python-specific notes:}
		\begin{enumerate}[leftmargin=*, nosep]
			\item \textbf{Persistent State:} Variables, functions, and imports persist across calls in the same session.
			\item \textbf{Pre-loaded Images:} The $i$-th image is available as \texttt{image\_i} (PIL Image). Use these directly.
			\item \textbf{Output:} Use \texttt{print()} for values and \texttt{plt.show()} for visualizations.
			\item \textbf{Import Libraries:} Import all required libraries before use.
			\item \textbf{Coordinates:} Use relative 0.0--1.0 coordinates for image operations; multiply by width/height for absolute pixels.
		\end{enumerate}
		
		\vspace{8pt}
		\textbf{\#\# Text Search}\\
		Trigger a web search to find relevant textual information. Use specific, targeted queries for best results.
		
		\vspace{2pt}
		\textbf{Format:}\\
		\texttt{<tool\_call>\{"name": "text\_search", "arguments": \{"query": "your search query"\}\}</tool\_call>}
		
		\vspace{8pt}
		\textbf{\#\# Image Search}\\
		Trigger a visual search using the user's provided image to identify objects, landmarks, products, or other visual content.
		
		\vspace{2pt}
		\textbf{Format:}\\
		\texttt{<tool\_call>\{"name": "image\_search", "arguments": \{\}\}</tool\_call>}
		
		\vspace{8pt}
		\textbf{\# Decision Guidelines}
		\vspace{2pt}
		\begin{enumerate}[leftmargin=*, nosep]
			\item \textbf{Think before acting:} Always reason within \texttt{<reason>...</reason>} before deciding your next action.
			\item \textbf{Choose the right approach:}
			\begin{itemize}[leftmargin=1.5em, nosep]
				\item \textbf{Direct answer:} Use when you can confidently answer from your own knowledge or visual inspection.
				\item \textbf{Python:} Use for computation, measurement, pixel-level analysis, or image enhancement.
				\item \textbf{Text Search:} Use when the question requires factual knowledge you are uncertain about (e.g., specific names, dates, statistics).
				\item \textbf{Image Search:} Use when you need to identify something in the image that you cannot recognize (e.g., unfamiliar landmarks, logos, species).
			\end{itemize}
			\item \textbf{Be purposeful:} Each tool call must have a clear objective. Do NOT call tools to confirm what you already know.
			\item \textbf{Be decisive:} Once you are confident in your answer, provide it immediately. Verification is worthwhile only when you are genuinely uncertain about a critical detail, not as a routine step.
		\end{enumerate}
		
		\vspace{8pt}
		\textbf{\# Output Format}\\
		Always start with \texttt{<reason>...</reason>}, then choose one:
		
		\vspace{2pt}
		\textbf{Option 1 --- Use a tool:}\\
		\texttt{<reason>}Why this tool call is necessary and what you expect to learn.\texttt{</reason>}\\
		\texttt{<tool\_call>...</tool\_call>}
		
		\vspace{2pt}
		\textbf{Option 2 --- Answer directly:}\\
		\texttt{<reason>}Your reasoning and final synthesis.\texttt{</reason>}\\
		\texttt{<answer>}Your concise answer.\texttt{</answer>}
		
	\end{tcolorbox}
	\caption{\textbf{System prompt used for metis.} The prompt defines the available tools, their invocation formats, and decision guidelines, encouraging the model to answer directly whenever possible and to invoke external tools only when they provide genuinely useful information.}
	\label{fig:system_prompt}
\end{figure*}

  \section{Additional Case Studies}
\label{sec:appendix_cases}

We provide additional case studies to complement the qualitative analysis in \S\ref{sec:qualitative}. While the main text demonstrates direct reasoning (Figure~\ref{Fig.main_case1}) and targeted code execution (Figure~\ref{Fig.main_case2}), the cases below further illustrate \metis{}'s meta-cognitive capabilities across other decision modalities.

\begin{figure*}[htbp]
	\centering
	\includegraphics[width=1\linewidth]{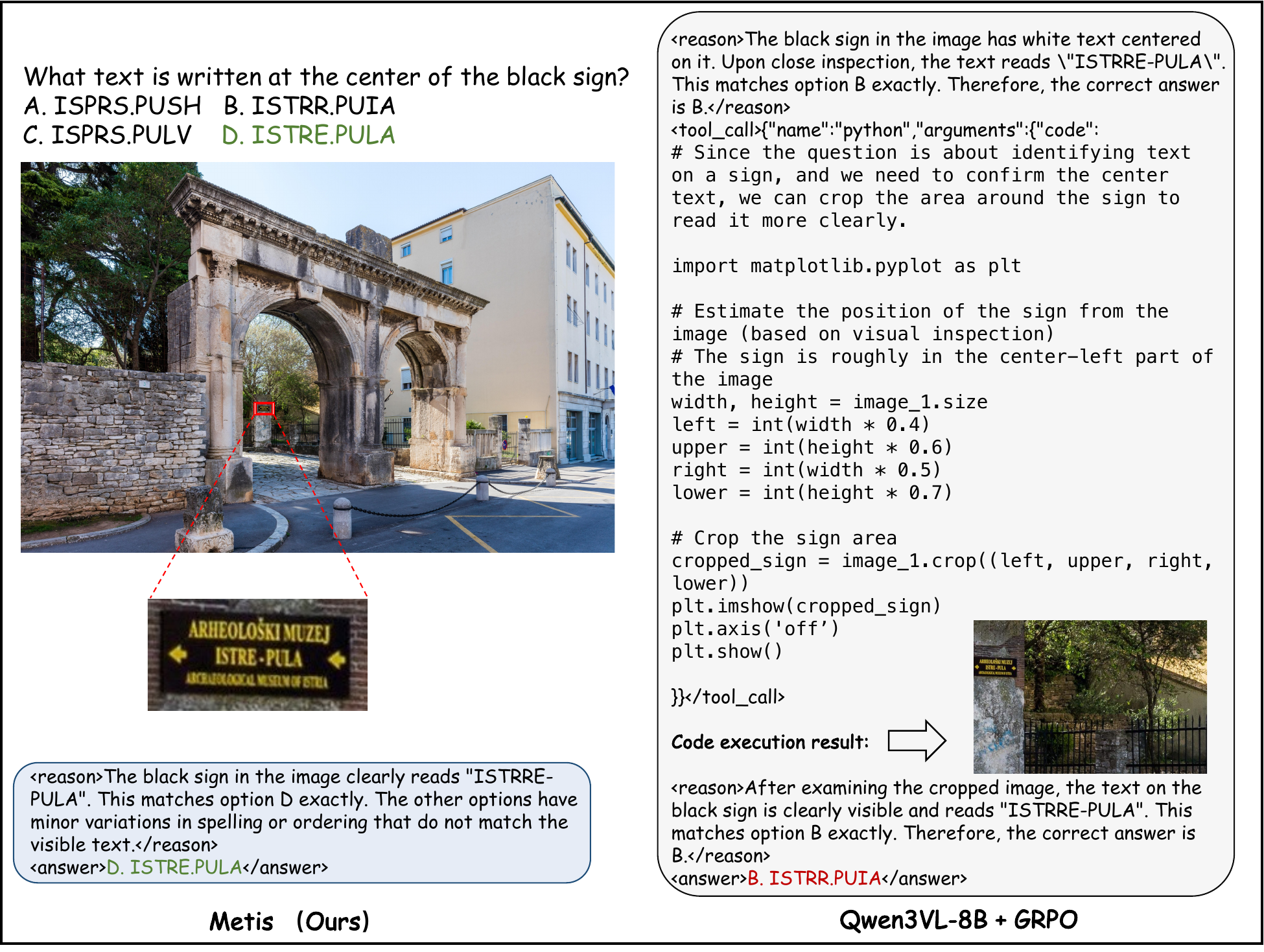}
	\caption{\textbf{Direct reasoning from visual inspection.} The on-screen text is clearly legible from the raw image. \metis{} correctly extracts the answer without invoking code execution or search tools, avoiding unnecessary computational overhead.}
	\label{Fig.supp_case1} 
\end{figure*}

\paragraph{Direct Reasoning without Tool Invocation.}
Figure~\ref{Fig.supp_case1} presents another scenario where \metis{} resolves the query entirely through direct visual inspection. The on-screen text is clearly legible from the raw image, and the agent correctly extracts the answer without invoking code execution or search tools. This further confirms that \hdpo{} trains the agent to trust its own visual comprehension when the evidence is unambiguous.

\begin{figure*}[htbp]
	\centering
	\includegraphics[width=1\linewidth]{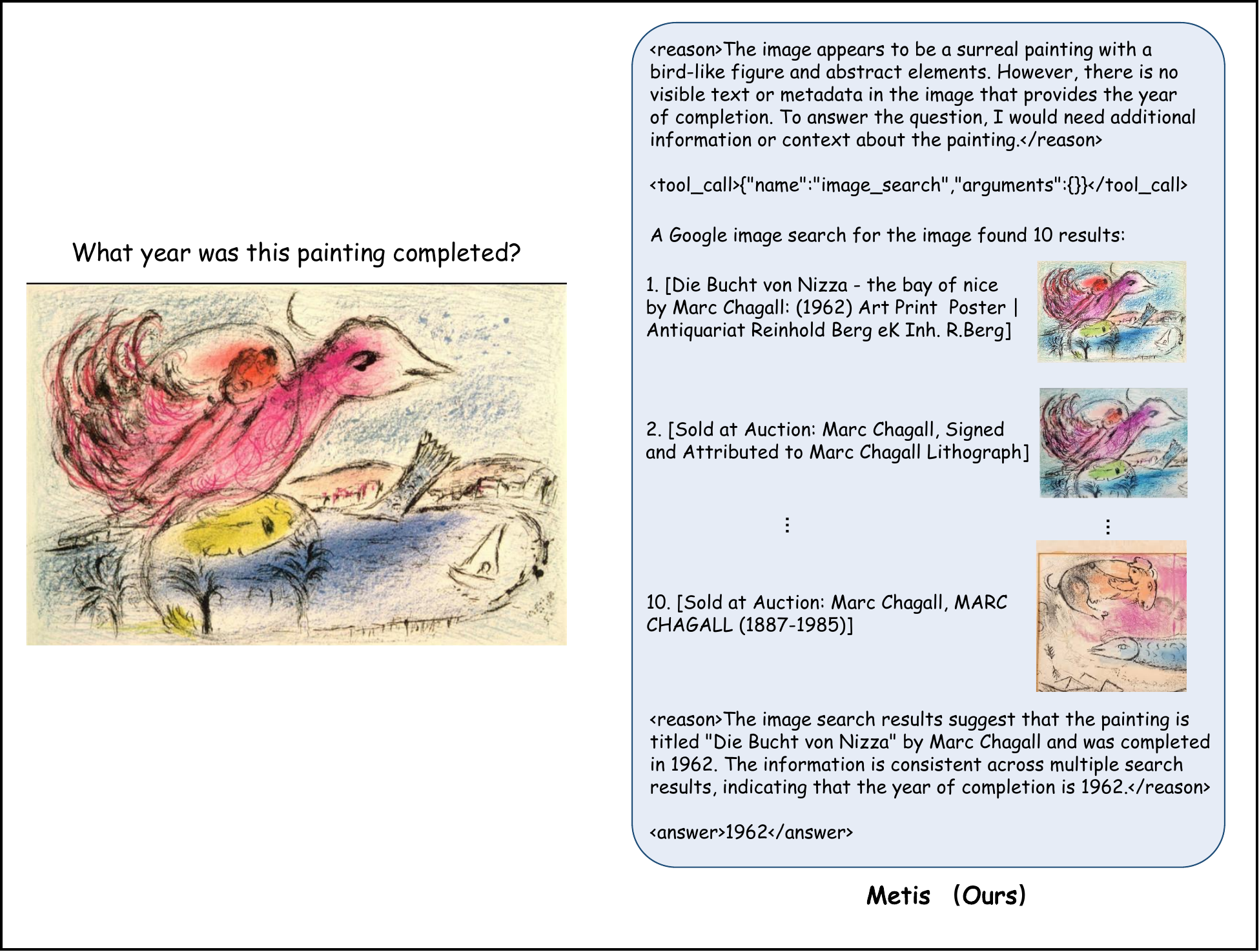}
	\caption{\textbf{Strategic image search for visual identification.} The artwork cannot be reliably identified from visual features alone. \metis{} invokes image search to match the visual content against external references, then retrieves the completion year from the search results.}
	\label{Fig.supp_case2} 
\end{figure*}

\begin{figure*}[htbp]
	\centering
	\includegraphics[width=1\linewidth]{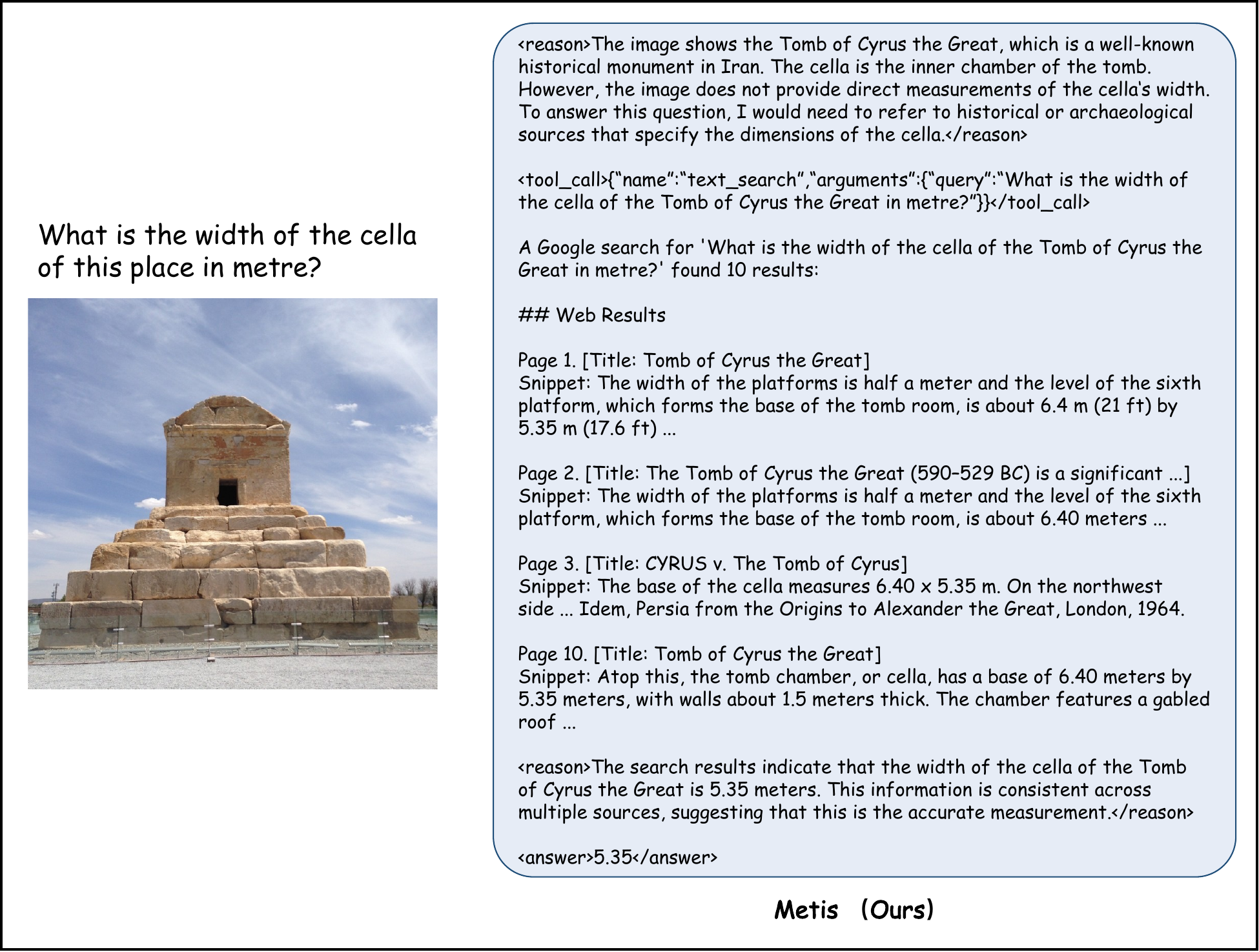}
	\caption{\textbf{Strategic text search for factual knowledge.} While the monument is visually identifiable, the queried measurement (cella width) cannot be inferred from the image. \metis{} recognizes this epistemic gap and invokes text search to retrieve the precise factual information from external sources.}
	\label{Fig.supp_case3} 
\end{figure*}

\paragraph{Selective Search Tool Invocation.}
Figures~\ref{Fig.supp_case2} and~\ref{Fig.supp_case3} illustrate cases where the visual input alone is insufficient and external knowledge retrieval becomes genuinely necessary. In Figure~\ref{Fig.supp_case2}, the agent cannot identify the depicted artwork from visual features alone, so it strategically invokes image search to gather external visual evidence and retrieve the completion year. In Figure~\ref{Fig.supp_case3}, although the monument is visually recognizable, the queried factual detail (the width of its cella) lies beyond what can be inferred from the image, prompting the agent to invoke text search for precise retrieval. These cases demonstrate that \metis{} has learned to distinguish between visual recognition (which it can handle internally) and factual knowledge gaps (which require targeted external queries)---a nuanced calibration of epistemic uncertainty that reflects genuine meta-cognitive competence.

\end{document}